\documentclass{article}

    \PassOptionsToPackage{numbers, compress}{natbib}
    \usepackage[preprint]{neurips_2022}

\usepackage[utf8]{inputenc} % allow utf-8 input
\usepackage[T1]{fontenc}    % use 8-bit T1 fonts
\usepackage{hyperref}       % hyperlinks
\usepackage{url}            % simple URL typesetting
\usepackage{booktabs}       % professional-quality tables
\usepackage{amsfonts}       % blackboard math symbols
\usepackage{nicefrac}       % compact symbols for 1/2, etc.
\usepackage{microtype}      % microtypography
\usepackage{graphicx}
\usepackage{multirow}
\usepackage{amsmath}
\usepackage{dcolumn}
\usepackage{multicol}
\usepackage{makecell}
\usepackage{siunitx}
\usepackage{wasysym}
\usepackage{xcolor}         % colors
\hypersetup{hidelinks}

\newcommand{\textarrow}[1]{\xrightarrow[]{#1}}
\renewcommand{\textleftarrow}[1]{\xleftarrow[]{#1}}

\title{MDMLP: Image Classification from Scratch on Small Datasets with MLP}

\author{
  Tian Lv \\
  Jiangsu University\\
  Zhenjiang, China \\
  \texttt{Amoza.light@gmail.com} \\
  \And
  Chongyang bai \\
  Dartmouth College \\
  Hanover, NH 03755 \\
  \texttt{bchy1023@gmail.com} \\
  \And
    Chaojie Wang \thanks{Corresponding author.} \\
    Jiangsu University\\
    Zhenjiang, China \\
  \texttt{cjwang@ujs.edu.cn} \\
}

\begin{document}

\maketitle

\begin{abstract}

  The attention mechanism has become a go-to technique for natural language processing and computer vision tasks. Recently, the MLP-Mixer and other MLP-based architectures, based simply on multi-layer perceptrons (MLPs), are also powerful compared to CNNs and attention techniques and raises a new research direction. 
  However, the high capability of the MLP-based networks severely relies on large volume of training data, and lacks of explanation ability compared to the Vision Transformer (ViT) or ConvNets. When trained on small datasets, they usually achieved inferior results than ConvNets.
  To resolve it, we present (i) multi-dimensional MLP (MDMLP), a conceptually simple and lightweight MLP-based architecture yet achieves SOTA when training from scratch on small-size datasets; (ii) multi-dimension MLP Attention Tool (MDAttnTool), a novel and efficient attention mechanism based on MLPs. Even without strong data augmentation, MDMLP achieves 90.90\% accuracy on CIFAR10 with only 0.3M parameters, while the well-known MLP-Mixer achieves 85.45\% with 17.1M parameters. In addition, the lightweight MDAttnTool highlights objects in images, indicating its explanation power. 
  Our code is available at \url{https://github.com/Amoza-Theodore/MDMLP}.
\end{abstract}

\section{Introduction}

\begin{figure}
    \centering
    \includegraphics[width=0.8\textwidth]{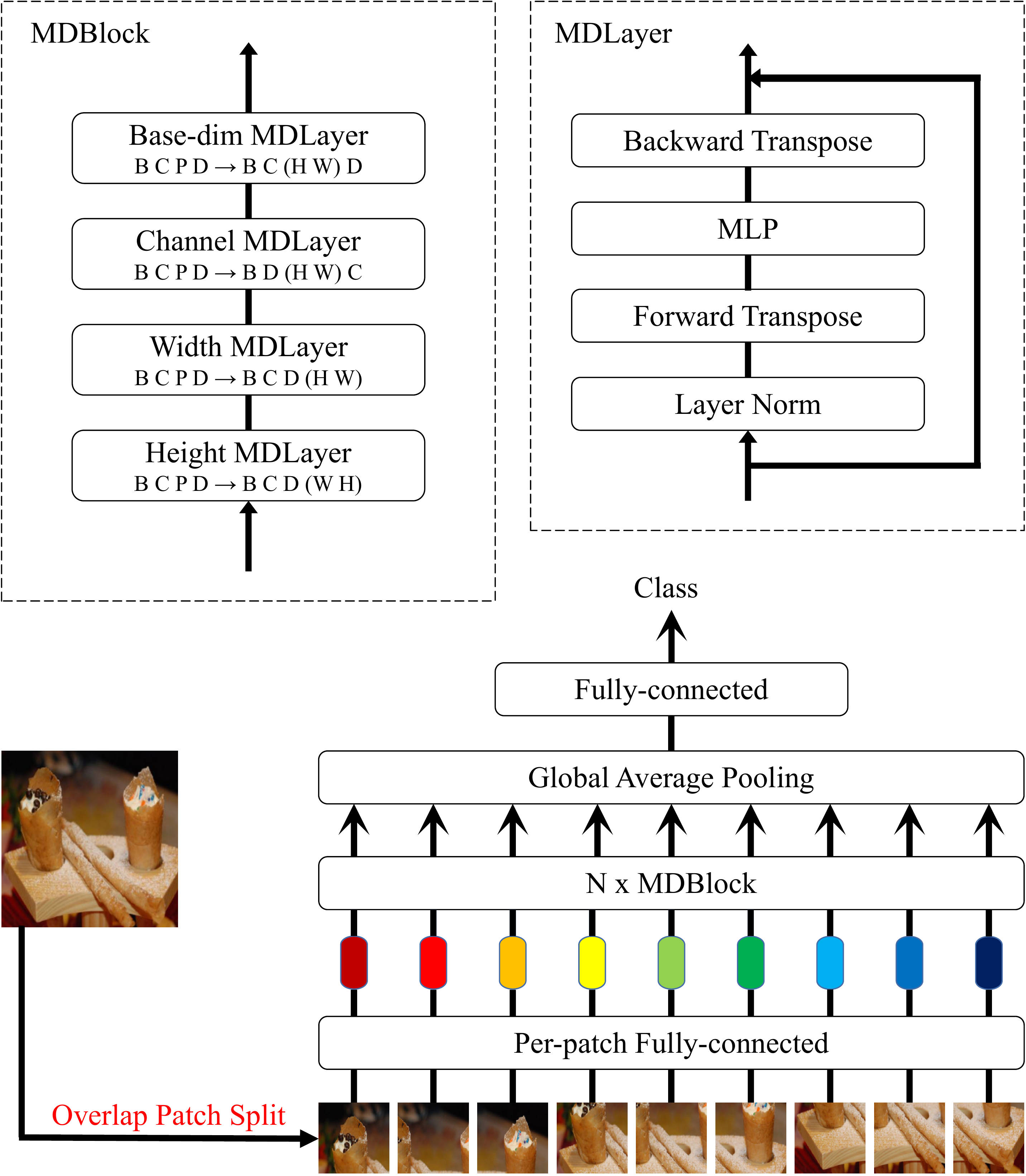}
    \caption{Model overview. MDMLP consists of overlapped per-patch linear embeddings, MDBlocks, and a classifier head. MDBlocks contain four MDLayers for height-mixing, width-mixing, channel-mixing and token-mixing (also referred as "base-dim"). MDLayer contains two linear layers and a GELU nonlinearity. Transpose in MDLayer means to rearrange tensors for making the last dimension to be the operating dimension. e.g. \ For Height MDLayer, we reshape "B C P D" to "B C D (W H)" so that the height dimension is put to the last dimension of the tensor.}
    \label{fig:arch}
\end{figure}

Transformer \cite{vaswani2017attention} with attention mechanism has become the dominant architecture in natural language processing (NLP). As a successful application of the attention mechanism to the computer vision (CV), Vision Transformer (ViT), and its variations \cite{dosovitskiy2020image, liu2021swin, touvron2021training, bao2021beit, wang2021pyramid}, has gained popularity in many fields. Lately, the MLP-Mixer model \cite{tolstikhin2021mlp}, which is purely based on multi-layer perceptrons (MLPs), has achieved a powerful performance compared to state-of-the-art architectures, demonstrating a new probable research trend. However, there are still many problems to be resolved.

First, MLP-Mixer and its variations \cite{tolstikhin2021mlp, hou2022vision, liu2021pay, tang2021image, lian2021mlp, touvron2021resmlp, chen2021cyclemlp} encode spatial information by flattening the spatial dimensions to the channel dimension (i.e. reshaped to "tokens$\times$channels"), resulting in the loss of positional and channel information. The Vision Permutator (ViP) \cite{hou2022vision}, which is closest to our work, 
splits image to non-overlapping patches and permutes along height, width, and channel to apply MLP. However, this loses relation and continuity information between patches. Imaging pixels around a split boundary are highly correlated and continuous, but the model has no clue about it after splitting.

Second, MLP-Mixer requires an immense amount of training dataset (14M-300M images), suffering the dilemma which cannot be trained in mid-size (1-10M images) or small-size datasets from scratch, and existing MLP models \cite{neyshabur2020towards, lin2015far, mocanu2018scalable, urban2016deep} have low accuracy for small-size datasets, compared to ConvNets. There has been a lot of work \cite{touvron2021resmlp, tang2021image, li2021convmlp} that attempts to apply the MLP architecture to mid-size datasets. For small-size datasets, however, only the vision transformer variant \cite{lee2021vision} achieves fairly good results compared to ConvNets like ResNet.
Although transfer learning is a good therapy for small-size datasets, on largely shifted domains, it may not be effective. 
It remains a problem whether MLP-based architecture can be trained well from scratch on a small-size dataset. 

Third, MLP-Mixer uses non-overlap patch embedding, which loses potential positional information. Much work \cite{trockman2022patches, yuan2021tokens, xiao2021early, yuan2021incorporating} has been done to explore this phenomenon. Our experiment has shown that the use of overlapped patches significantly improves performance by approximately 3\%.

Fourth, MLP-Mixer lacks an effective means of explanation. For ViT, attention visualization and interpretability \cite{chefer2021transformer} are important factors in its success, but for MLP-Mixer and similar methods, it's unclear what are learned from training. EAMLP \cite{guo2021beyond} uses linear layers to realize the multi-attention mechanism, but this is not a general approach (as applied to ConvNets) and is still limited to the self-attention mechanism.

In view of the above problems, we propose an efficient architecture called multi-dim MLP (MDMLP). It accepts a sequence of linearly projected image patches (i.e. tokens) with overlap as input, keeping its dimensionality for not losing any information. MDMLP uses four types of MLP layers: height-mixing, width-mixing, channel-mixing, and token-mixing MLPs with normalization. In addition, for visualization, we propose a multi-dim MLP Attention Tool (MDAttnTool). It consists of two layers of MLPs with eight hidden units and an average pooling layer to get weights.

Despite its simplicity, MDMLP with MDAttnTool has achieved excellent performance. Without strong augmentations, MDMLP achieves 90.90\% on CIFAR10 along with 0.30M parameters and 0.28G flops, while MLP-Mixer \cite{lian2021mlp} can only achieve 85.45\% together with 17.1M parameters and 1.21G flops, and Neyshabur's model \cite{neyshabur2020towards} only achieves 85.19\%. MDAttnTool successfully visualizes the attention area as well as the object boundary not only on MDMLP but also on ResNet \cite{he2016deep}, which proved to be a general visualization method, breaking through the existing self-attention mechanism.

In summary, our contributions are as follows:
\begin{itemize}
    \item We propose multi-dim MLP (MDMLP) to train from scratch on small-size datasets and gain the highest accuracy, with yet extremely low parameters and flops which are comparable to ConvNets.
    \item We propose overlap patch embedding to make up for the loss of adjacent information caused by the split of neighboring image blocks.
    \item We propose MLP Attention Tool (MLPAttnTool) to visualize the feature learned by the neural network, which is a general tool applicable not only to our model but also to other models such as ConvNets.
\end{itemize}

\section{Related Works}
Much of the work has concentrated on pure MLP architectures on small-size datasets such as CIFAR10 \cite{krizhevsky2009learning}, but not nearly as accurate as ours. Various works \cite{mocanu2018scalable, urban2016deep} are committed to making MLP competitive with state-of-the-art models. \citet{lin2015far} used fully connected networks together with heavy data augmentation and pre-training, achieving 70\% accuracy on CIFAR10. \citet{tolstikhin2021mlp} worked out the MLP-Mixer, attaining approximately 80\% accuracy on CIFAR10 if trained from scratch. \citet{neyshabur2020towards} developed an MLP architecture with a regularizer constraining the model to be closed to networks based on conv and get 85.19\% accuracy on CIFAR10. Our work, MDMLP, conceptually simple and extremely light in the terms of parameters and flops but reaches high accuracy without heavy augmentations (mixup, cutmix, etc.).

Pure MLP architectures have thrived on large-size datasets (14M-300M images, such as JFT-300M \cite{sun2017revisiting}) and mid-size datasets (1-10M images, such as IMAGENET-1K \cite{deng2009imagenet}) with heavy augmentations, but have yet to overcome small datasets. The MLP-Mixer \cite{tolstikhin2021mlp} successfully achieves state-of-the-art performance on large-size datasets. Plenty of works \cite{touvron2021resmlp, liu2021pay, hou2022vision, tang2021image, hou2022vision, ding2021repmlp} are successful with mid-size datasets in Image Classification. Other work \cite{lian2021mlp, chen2021cyclemlp, guo2021hire} demonstrates that the pure MLP architecture can also gain fantastic performance in Object Detection and Semantic Segmentation. \citet{hou2022vision} brought up the Vision Permutator with height and width dimensions, much similar to our work which keeping height, width and channel dimensions. By comparison, Our work focus on small-size datasets without heavy augmentations, CIFAR10, CIFAR100, Flowers102, and so on.

The work of attention mechanism \cite{vaswani2017attention} in computer vision \cite{dosovitskiy2020image} makes "Patch Embed" a core module in model design field. \citet{cordonnier2019relationship} extracted 2 $\times$ 2 patches from the input image (referred as "Patch Embed") and applied self-attention layers on CIFAR10. Also, a lot of ViT variants \cite{liu2021swin, touvron2021training, bao2021beit, wang2021pyramid} prove the excellent performance of ViT on large-size and mid-size image classification datasets, Object Detection and Semantic Segmentation. Other work \cite{he2021masked, he2020momentum} center on Unsupervised learning and transfer learning. \citet{lee2021vision} put forward shift layer to make ViT success in small-size datasets. \citet{yu2021metaformer} advised that the attention component is not necessary in transformer block. Our work use overlap patch embedding (i.e.\ split patch with overlap pixels) greatly improving our model's performance.

Many have attempted to make attention mechanism and MLP architecture explainable. Vision transformer \cite{lee2021vision} itself has visualized attention layer and positional embeddings. \citet{chefer2021transformer} visualize attention layer in an image more clearly. \citet{hao2020self} visualize attention layer in words and explain the inner properties. \citet{touvron2021resmlp} visualize the MLP's positional embeddings. In our work, we use a 2-layer fully connected network with 8 hidden units as a new attention add-on to visualize the MLP's learned representations, successfully draw the attention area and the object boundary.

The model design field is full of vitality, to some degree, by modern training strategies and data augmentations methods. \citet{he2016deep} developed residual connection, to handle the problem of gradient disappearance. \citet{ba2016layer} raised layer norm, making it possible to norm data of different shapes. Moreover, \cite{wightman2021resnet} and \citet{liu2022convnet} use modern model design approaches, making the out-of-date convnet comes alive again. The Pytorch \cite{paszke2017automatic}, timm \cite{rw2019timm} and einops library \cite{einops} provide a lot of add-ons for data augmentations and processing such as mixup \cite{zhang2017mixup}, cutmix \cite{yun2019cutmix}, label smoothing \cite{muller2019does}, random erase \cite{zhong2020random}, EMA \cite{klinker2011exponential}, stochastic depth \cite{huang2016deep}, AutoAugment \cite{cubuk2018autoaugment}, RandAugment \cite{cubuk2020randaugment}. But our work concentrate on lightweight the model, we do not use strong augmentation in our training regime in terms of this.

\section{MDMLP Architecture}

% not sharing parameters
Figure \ref{fig:arch} shows the overall architecture of MDMLP. In a nutshell, MDMLP accepts a sequence of overlapped patch embeddings maintaining the height, width, channels and base-dim dimensions without losing any information. Then, it applies N layers of MDBlocks, each of which consists of FC layers applied to each dimension of those patches in turn. Last, it utilizes a standard classification head, global average pooling and linear classifier.

\subsection{Overlap Patch Embedding}

% some other related works
ViTs \cite{lee2021vision} has introduced Patch Embed, which splits images into patches followed by a linear layer to map patches to tokens, but without overlap. 
Essentially, non-overlap patch embedding can be replaced by a convolution operation where stride is equal to kernel size. But for traditional Convolutional Neural Networks (CNNs), it is common that stride is smaller than kernel size, i.e.\ with overlap. Inspired by this, in MDMLP, we also use overlap patch embedding to obtain the neighboring information between connected patches.

The calculation of overlap patch embedding is the same as the convolution operation.

\begin{equation*}
\begin{aligned}
& \mathbb{R}^{C\times H\times W}\textarrow{\text{Overlap Patch Split}}\mathbb{R}^{H'\times W'\times C\times P}\\
& \mathbb{R}^{H'\times W'\times C\times P}\textarrow{\text{Patch Embedding}}\mathbb{R}^{H'\times W'\times C\times D}
\end{aligned}
\end{equation*}
\begin{equation}
\begin{aligned}
& P = p ^ 2 \\
& H' = (H - p) / O + 1 \\
& W' = (W - p) / O + 1
\end{aligned}
\end{equation}

where the second mapping is a linear embedding layer, $P$ is the patch size, $O$ is the overlap size.

The overlap patch embedding, to some extent, provides not only positional information but also a larger receptive field.

\subsection{MDBlock}

Vision Permutator \cite{hou2022vision} keeps the dimension of height, width and channel (hwc), and MDMLP maintains the dimension of height, width, channels and base-dim (also referred as "token") dimensions (hwcd). The advantage of this is not losing any dimensional information and saving memory to a large extent.

Multi-dim Block (MDBlock) consists of four Multi-dim Layers (MDLayers) mixing height, width, channel and base-dim (token) information.When an image has been mapped to overlapped patch embeddings in space $\mathbb{R}^{H'\times W'\times C\times D}$, we then apply N $\times$ Blocks following a Global Average Pooling layer to get the result of class.

\subsection{MDLayer}
As the core component of MDMLP, it is composed of transpose and linear layers. Take the Height MDLayer for example, we firstly apply a Layernorm, then arrange (also referred as "transpose") it to $\mathbb{R}^{D\times W'\times C\times H'}$ followed by an MLP add-on which consists of 2 layers of MLP. We set the hidden units to $H' \times f$ where $f$ is called expansion factor. GELU activation and dropout have been employed in the MLP layer. Afterwards, we rearrange it to the original shape $\mathbb{R}^{H'\times W'\times C\times D}$ with a self-learning residual connection.

% The whole process is as follows. When an image has been mapped to overlapped patch embeddings in space $\mathbb{R}^{H'\times W'\times C\times D}$, we firstly apply a Layernorm, then arrange (also referred as "transpose") it to $\mathbb{R}^{D\times W'\times C\times H'}$ followed by a MLP add-on which consists of 2 layers of MLP. We set the hidden units to $H' \times f$ where $f$ is called expansion factor. GELU activation and dropout have been employed in the MLP layer. Afterwards, arrange it to $\mathbb{R}^{D\times H'\times C\times W'}$ followed by a MLP add-on. The channel and base-dim dimension are as the same. When all 4 dimension operations has been done, then we rearrange it to the original shape $\mathbb{R}^{H'\times W'\times C\times D}$ with a self-learning residual connection.

MLP can be formulated as follows (not showing bias and dropout):
\begin{equation}
\begin{aligned}
% no dropout
& MLP(X) =  X +  \sigma(W_2\sigma (W_1X)) \\
& \text{where } W_1 \in \mathbb{R}^{(f\times n)\times n}, W_2 \in \mathbb{R}^{n\times (f\times n)}
\label{form:mlp}
\end{aligned}
\end{equation}

MDLayer can be expressed as follows (Height MDLayer for example):
\begin{equation}
\begin{aligned}
X_0 & = LayerNorm(X) \\
X_0 & \in \mathbb{R}^{H'\times W'\times C\times D}\textarrow{\text{transpose}}X_0' \in \mathbb{R}^{D\times W'\times C\times H'} \\
Y_0 & = MLP(X_0'), X_0' \in \mathbb{R}^{D\times W'\times C\times H'} \\
Y_0 & \in \mathbb{R}^{D\times W'\times C\times H'}\textarrow{\text{transpose}}Y_0' \in \mathbb{R}^{H'\times W'\times C\times D} \\
Y & \textleftarrow{\text{residual}}  Y_0' +  X, X \in \mathbb{R}^{H'\times W'\times C\times D}
\end{aligned}
\end{equation}

% \begin{equation}
% \begin{aligned}
% X & \in \mathbb{R}^{H'\times W'\times C\times D}\textarrow{transpose}X' \in \mathbb{R}^{D\times W'\times C\times H'} \\
% Y_1 & = MLP_{h'}(X'), X' \in \mathbb{R}^{D\times W'\times C\times H'} \\
% Y_1 & \in \mathbb{R}^{D\times W'\times C\times H'}\textarrow{transpose}Y_1' \in \mathbb{R}^{D\times H'\times C\times W'} \\
% Y_2 & = MLP_{w'}(Y_1'), Y_1' \in \mathbb{R}^{D\times H'\times C\times W'} \\
% Y_2 & \in \mathbb{R}^{D\times H'\times C\times W'}\textarrow{transpose}Y_2' \in \mathbb{R}^{D\times H'\times W'\times C} \\
% Y_3 & = MLP_{c}(Y_2'), Y_2' \in \mathbb{R}^{D\times H'\times W'\times C} \\
% Y_3 & \in \mathbb{R}^{D\times H'\times W'\times C}\textarrow{transpose}Y_3' \in \mathbb{R}^{C\times H'\times W'\times D} \\
% Y_4 & = MLP_{d}(Y_3'), Y_3' \in \mathbb{R}^{C\times H'\times W'\times D} \\
% Y_4 & \in \mathbb{R}^{C\times H'\times W'\times D}\textarrow{transpose}Y_4' \in \mathbb{R}^{H'\times W'\times C\times D} \\
% Y & \textleftarrow{residual}  Y_4' +  X, X \in \mathbb{R}^{H'\times W'\times C\times D}
% \end{aligned}
% \end{equation}

\section{Experiments}

In this section, several experiments are conducted to verify that the proposed method improves the performance of MLP architecture in terms of parameters, flops and accuracy.
Sec. \ref{sec:Settings} describes the experiment setup. Section \ref{sec:Quantitative Result} shows quantitatively that the proposed method effectively improves the MLP architecture and achieves performance comparable to CNN.

\subsection{Settings}
\label{sec:Settings}

\subsubsection{Environment and Dataset}
\label{sec:Environment and dataset}
The proposed method was implemented in Pytorch \cite{paszke2017automatic}. CIFAR-10, CIFAR-100 \cite{krizhevsky2009learning}, Flowers-102 \cite{nilsback2008automated} and Food101 \cite{bossard14} were employed in our experiments whose image size was resized to 224, and the GPU was V100 \cite{nvidia2017nvidia}.

\subsubsection{Model Configurations}
In the case of MDMLP, the configuration was determined experimentally. The hidden dimension was set to 64, the depth was set to 8, the expansion factor ($f$) was set to 4. The patch size was set to 4, the overlap size was set to 2 on CIFAR-10 and CIFAR-100 datasets, while the patch size was set to 14, the overlap size was set to 7 on Flowers102 and Food101 datasets.

For ResNet20, we use ResNet \cite{he2016deep} without making any changes. It is worth mentioning that we adopt the implementation of \citet{Idelbayev18a} instead of timm \cite{rw2019timm} whose parameters and flops are one hundred times greater than the former.

In other cases, the tiny model configurations presented in the corresponding papers were adopted as they were respectively considering the trade-off among parameters, flops and accuracy except that the patch size was determined experimentally. For ViT \cite{dosovitskiy2020image}, the hidden dimension was set to 192, mlp ratio to 2, depth to 9, heads to 12, patch size to 4 on CIFAR-10 and CIFAR-100 as well as 14 on Flowers102 and Food101. For AS-MLP \cite{lian2021mlp}, embed dimmension to 96, depth to [2, 2, 6, 2], shift size to 5, patch size to 2 on CIFAR-10 and CIFAR-100 as well as 7 on Flowers102 and Food101. For gMLP \cite{liu2021pay}, number of blocks to 30, embed dimension to 128, mlp ratio to 6, patch size to 4 on CIFAR-10 and CIFAR-100 as well as 14 on Flowers102 and Food101. For ResMLP \cite{touvron2021resmlp}, number of blocks to 12, embed dimension to 384, mlp ratio to 4, patch size to 4 on CIFAR-10 and CIFAR-100 as well as 28 on Flowers102 and Food101. For ViP \cite{hou2022vision}, number of layers to [4, 3, 8, 3], mlp ratio to 3, embed dimension to 384, the patch size to 4 on CIFAR-10 and CIFAR-100 as well as 28 on Flowers102 and Food101. For MLP-Mixer \cite{tolstikhin2021mlp}, number of blocks to 8, embed dimension to 512, patch size to 4 on CIFAR-10 and CIFAR-100 as well as 14 on Flowers102 and Food101. For S-FC ($\beta$-LASSO), the result was from the corresponding paper \cite{neyshabur2020towards}.

\subsubsection{Training Regime}

\textit{We do not use any strong augmentations or regularization techniques in all training configurations contrary to other papers.} i.e. \ cutmix, auto augment, label smoothing, etc. What we only use are horizontal flip and color jitter. 

The training strategy was adopted from ResNet \cite{he2016deep} without any changes based on experiments. The SGD \cite{shamir2013stochastic} was used as the optimizer. Weight decay was set to 0.0001, batch size to 128 (however, 16 for flowers102), and warm-up to 10. All models were trained for 200 epochs, the cosine learning rate decay was used, and the initial learning rate was set to 0.1. All the training was done on timm \cite{rw2019timm}. 

\subsection{Quantitative Result}
\label{sec:Quantitative Result}

\subsubsection{Image Classification}

\begin{table}
    \caption{Experiments on CIFAR-10 and CIFAR-100}
    \label{tab:exp1}
    \centering
    \begin{tabular}{*{8}{c}}
        \toprule
        
        \multirow{2}{*}{FAMILY} & \multirow{2}{*}{MODEL} & \multirow{2}{*}{PARAMS (M)} & \multirow{2}{*}{FLOPS (G)} & \multicolumn{2}{c}{ACC (\%)} \\ \cmidrule(lr){5-6}
        & & & & \small{CIFAR10} & \small{CIFAR100} \\ \midrule
        
        Conv & ResNet20 & 0.27 & 0.04 & 91.99 & 67.39 \\
        % Conv & ResNet56 & 0.85 & 0.13 & 93.49 & 70.86 \\
        % Conv & convmixer & 0.71 & 0.56 & 94.05 & 73.51 \\ 
        Trans & ViT & 2.69 & 0.19 & 86.57 & 60.43 \\ \cmidrule(lr){1-6}
        MLP & AS-MLP & 26.20 & 0.33 & 87.30 & 65.16 \\
        % MLP & cyclemlp & 14.64 & 0.29 & 91.45 \\
        MLP & gMLP & 4.61 & 0.34 & 86.79 & 61.60 \\
        MLP & ResMLP & 14.30 & 0.93 & 86.52 & 61.40 \\
        MLP & ViP & 29.30 & 1.17 & 88.97 & \textbf{70.51} \\
        % MLP & wavemlp & 16.68 & 0.20 & 92.06 & \\
        MLP & MLP-Mixer & 17.10 & 1.21 & 85.45 & 55.06 \\ \cmidrule(lr){1-6}
        MLP & \small{S-FC ($\beta$-LASSO)} & - & - & 85.19 & 59.56 \\ \cmidrule(lr){1-6}
        MLP & MDMLP (ours) \label{mdmlp1} & 0.30 & 0.28 & 90.90 & 64.22 \\
        % MLP & mdmlp\label{mdmlp2} & \textbf{0.54} & \textbf{0.09} & 89.64 & 64.04 \\
        \bottomrule
    \end{tabular}
    
    \medskip{}\medskip{}
    
    \caption{Experiments on FLOWERS102 and FOOD101}
    \label{tab:exp2}
    \centering
    \begin{tabular}{*{8}{c}}
        \toprule
    
        \multirow{2}{*}{FAMILY} & \multirow{2}{*}{MODEL} & \multirow{2}{*}{PARAMS (M)} & \multirow{2}{*}{FLOPS (G)} & \multicolumn{2}{c}{ACC (\%)} \\ \cmidrule(lr){5-6}
        & & & & \small{FLOWERS102} & \small{FOOD101} \\ \midrule
        
        Conv & ResNet20 & 0.28 & 2.03 & 57.94 & 74.91 \\
        % Conv & ResNet56 & 0.86 & 6.28 & 58.43 & 81.90 \\
        % Conv & convmixer & 0.77 & 0.60 & 72.64 &  \\ 
        % Conv & convmixer & 0.88 & 0.18 & 65.00 &  \\ 
        Trans & ViT & 2.85 & 0.94 & 50.69 & 66.41 \\ \cmidrule(lr){1-6}
        MLP & AS-MLP & 26.30 & 1.33 & 48.92 & 74.92 \\
        MLP & gMLP & 6.54 & 1.93 & 47.35 & 73.56 \\
        MLP & ResMLP & 14.99 & 1.23 & 45.00 & 68.40 \\
        MLP & ViP & 30.22 & 1.76 & 42.16 & 69.91 \\
        MLP & MLP-Mixer & 18.20 & 4.92 & 49.41 & 61.86 \\ \cmidrule(lr){1-6}
        MLP & MDMLP (ours) & 0.41 & 1.59 & \textbf{60.39} & \textbf{77.85} \\ 
        
        \bottomrule
    \end{tabular}

\end{table}

This section presents the experimental results for small-size datasets.

As showed in Table \ref{tab:exp1} On CIFAR10 and CIFAR100 datasets, our model achieves the significant reduction of parameters and flops under the competitive accuracy. MDMLP achieves the best accuracy of 90.90\% among all MLP and Transformer model families on CIFAR10 with the 0.30M parameters which is at least ten times smaller than others, achieving the comparable performance with ConvNets in terms of parameters, flops and accuracy.

Moreover, on flowers102 and food101 datasets (Table \ref{tab:exp2}), our model beats all other models in all three metrics ways. MDMLP achieves 60.39\% and 77.85\% respectively on flowers102 and food101, with 0.41M parameters and 1.59G flops.

\subsubsection{Ablation Study}

\begin{table}
    \centering
    \caption{Ablation Study on CIFAR-10 and CIFAR-100}
    \label{tab:ablation}
    % \begin{tabular}{ccccc}
    %     \toprule
    %     \multirow{2}{*}{MODEL} & \multirow{2}{*}{PARAMS (M)} & \multirow{2}{*}{FLOPS (G)} & \multicolumn{2}{c}{ACC (\%)} \\ \cmidrule(lr){4-5}
    %     & & & CIFAR10 & CIFAR100 \\ \midrule
    %     mdmlp (ours) & 0.30 & 0.28 & \textbf{90.90} & \textbf{64.22} \\
    %     -- overlap & 0.31 & 0.32 & 88.87 & 60.95 \\
    %     -- multi-dim & 17.8 & 4.26 & 87.09 & 57.08 \\
    %     -- overlap, multi-dim (MLP-Mixer) & 17.1 & 1.21 & 85.45 & 55.06 \\
    % \bottomrule
    % \end{tabular}
    \begin{tabular}{ccc}
        \toprule
        \multirow{2}{*}{MODEL} &  \multicolumn{2}{c}{ACC (\%)} \\ \cmidrule(lr){2-3}
        & CIFAR10 & CIFAR100 \\ \midrule
        MDMLP (ours) & \textbf{90.90} & \textbf{64.22} \\
        -- overlap & 88.87 & 60.95 \\
        -- mdblock & 87.09 & 57.08 \\
        -- overlap, mdblock (MLP-Mixer) & 85.45 & 55.06 \\
    \bottomrule
    \end{tabular}
\end{table}

This section shows that both our MDBlock and overlap patch embedding method play significant roles in the MDMLP model.

On one hand, the overlap patch embedding method, which provides more information between patches, improves the accuracy by 2\% and 4\% respectively on cifar10 and cifar100 datasets. We attribute this to the elevation of the receptive field.

On the other hand, the MDBlock architecture contributes to the whole performance by 3.8\% on CIFAR10 and 7.1\% on CIFAR100. Further, if both MDBlock and overlap patch embedding components are removed, our MDMLP model is essentially MLP-Mixer and the prediction accuracy drops again. 

\section{MDAttnTool for Visualization}

\begin{figure}
    \centering
    
    \begin{minipage}[t]{0.48\linewidth}
        \flushright
        \includegraphics[width=0.4\linewidth]{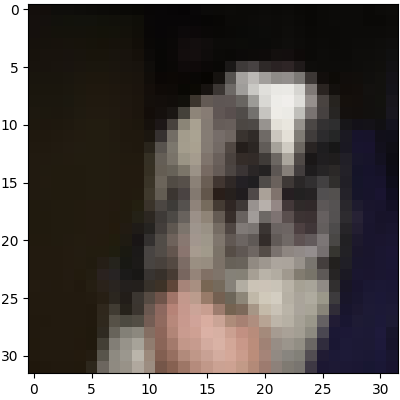}
    \end{minipage}
    \begin{minipage}[t]{0.48\linewidth}
        \flushleft
        \includegraphics[width=0.4\linewidth]{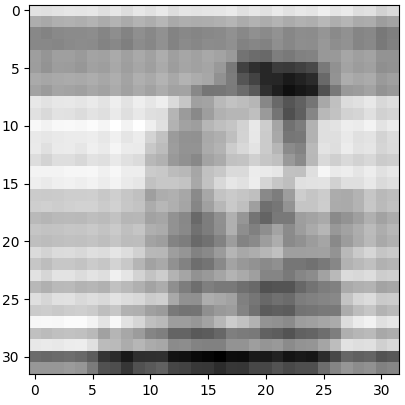}
    \end{minipage}
    
    \begin{minipage}[t]{0.48\linewidth}
        \flushright
        \includegraphics[width=0.4\linewidth]{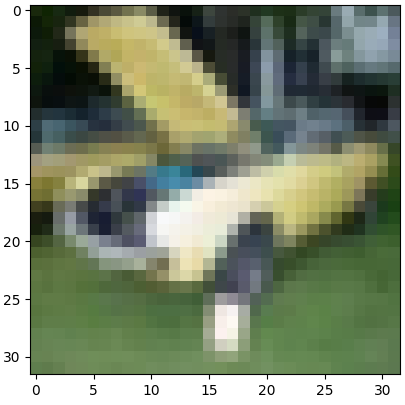}
    \end{minipage}
    \begin{minipage}[t]{0.48\linewidth}
        \flushleft
        \includegraphics[width=0.4\linewidth]{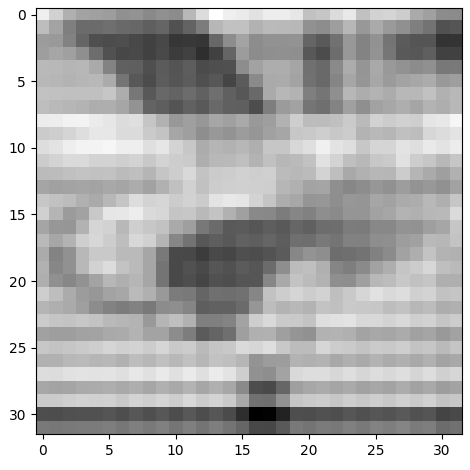}
    \end{minipage}
    
    \begin{minipage}[t]{0.48\linewidth}
        \flushright
        \includegraphics[width=0.4\linewidth]{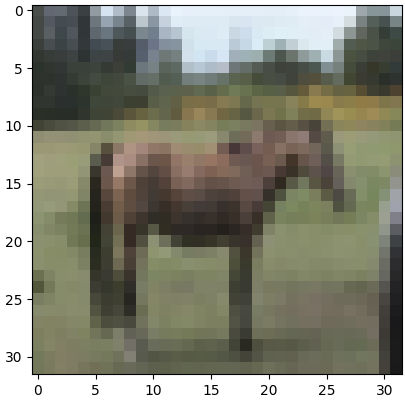}
    \end{minipage}
    \begin{minipage}[t]{0.48\linewidth}
        \flushleft
        \includegraphics[width=0.4\linewidth]{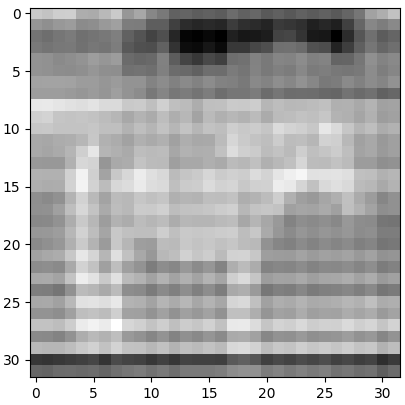}
    \end{minipage}
    
    \caption{MDAttnTool Image Visualization.
    The absolute values of image attention weights have been shown in gray mode on CIFAR10 validation set. Weights are initialized to 1 at the begining.}
    \label{fig:visual}
\end{figure}

\begin{figure}
    \centering
    \includegraphics[width=0.4\linewidth]{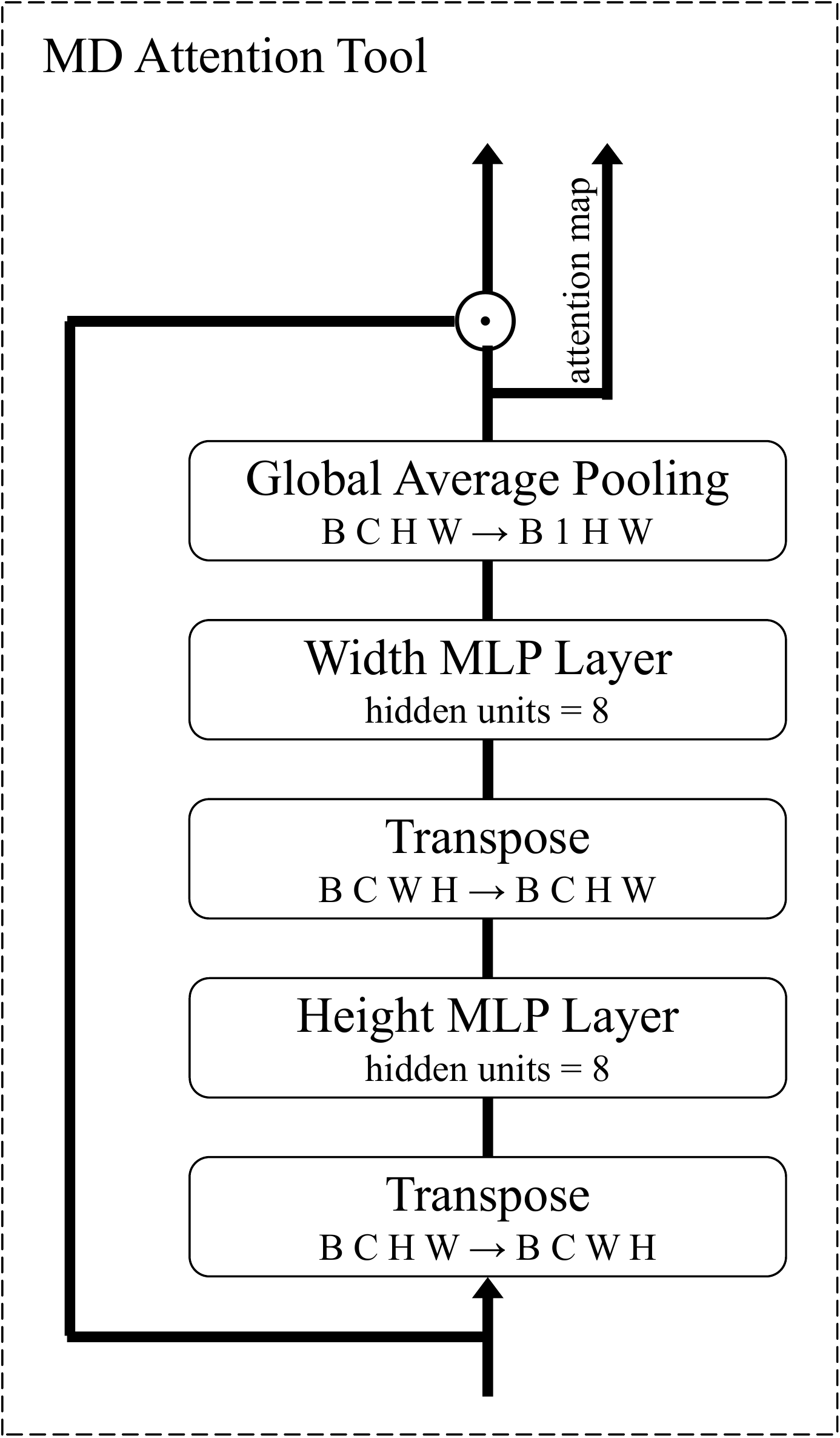}
    \caption{MDAttnTool. The input was transposed to "B C H W" followed by an MLP Layer and then transposed to "B C W H" to get the Height features. Then with a Global Average Pooling getting the weights which will be elementwise produced to the input.}
    \label{fig:MDAttnTool}
\end{figure}

ViT and its variants \cite{lee2021vision} has shown the attention values in visualization attributed to the attention layer, so we arise an MLP based attention layer for visualization. The Multi-dim Attention Layer (MDAttnTool) consists of 2 layers where the number of hidden units is 8, as the same to MLP layer (formula \ref{form:mlp}) but different in Global Average Pooling layer.

\noindent \textbf{MLP in MDAttnTool can be written as follows. (not shown bias and dropout)}
\begin{equation}
\begin{aligned}
& MLP(X) = X + \sigma(W_2\sigma (W_1Norm(X))) \\
& W_1 \in \mathbb{R}^{m\times n}, W_2 \in \mathbb{R}^{n\times m}
\end{aligned}
\end{equation}

\noindent \textbf{MDAttnTool for Image can be written as follows.}
\begin{equation}
\begin{aligned}
Y_1 & = MLP_w(X), X \in \mathbb{R}^{C\times H\times W} \\
Y_1 & \in \mathbb{R}^{C\times H\times W}\textarrow{transpose}Y_1' \in \mathbb{R}^{C\times W\times H} \\
Y_2 & = MLP_h(Y_1'), Y_1' \in \mathbb{R}^{C\times W\times H} \\
Y_2 & \in \mathbb{R}^{C\times W\times H}\textarrow{transpose}Y_2' \in \mathbb{R}^{C\times H\times W} \\
V & = AvgPool(Y_2'), \mathbb{R}^{C\times H\times W}\textarrow{transpose}\mathbb{R}^{1\times H\times W} \\
Y & = V \odot X
\end{aligned}
\end{equation}

where AvgPool is referred to "Global Average Pooling", and V (also referred to as "weights") are initialized to 1.0 at the beginning.

Multi-dim MLP Attention Tool has successfully shown the learned features in Figure \ref{fig:visual}, which contains not only attention but also some positional information (shown as object boundary). By the way, the pure MLP based attention architecture is extremely light compared to Vision Transformer \cite{lee2021vision}.

In addition, MDAttnTool breaks through the existing self-attention mechanism compared to EAMLP \cite{guo2021beyond}. Though EAMLP uses pure MLP architecture to get the attention, our method use elementwise product replacing the self-attention mechanism.
\section{Conclusion and Future Work}
\label{sec:Conclusion and Future}

To explore whether MLP-based architecture can achieve great visual recognition results when training on small datasets from scratch just as ConvNets, we propose MDMLP, which takes up to ten times smaller parameters than the existing MLP models. To visualize the model, we propose MDAttnTool, which is a general method to highlight the attention of models. However, one limitation is that the throughput is not as expected compared to our good-performing FLOPs. Although the current model still cannot outperform SOTA results on all datasets, we hope this work open a door to more revisits of MLP architectures in small-scale datasets for visual recognition tasks.

% \begin{table}[tp]
%     \caption{Ablation experiments on flowers102 and food101}
%     \label{tab:ablation}
%     \centering
%     \begin{tabular}{lcccc}
%         \toprule
%         \multirow{2}{*}{MODEL} & \multirow{2}{*}{PARAMS (M)} & \multirow{2}{*}{FLOPS (G)} & \multicolumn{2}{c}{ACC (\%)} \\ \cmidrule(lr){4-5}
%         & & & FLOWERS102 & FOOD101 \\ \midrule
%         baseline & 0.41 & 1.59 & 59.31 &  \\
%         -- overlap &  &  &  &  \\
%         -- multi-dim & 21.12 & 18.47 & 51.76 &  \\
%         % MLP-Mixer & 17.1 & 1.21 & 85.45 \\
%     \bottomrule
%     \end{tabular}
% \end{table}

\medskip

{
\small
\bibliographystyle{unsrtnat}
\bibliography{main.bib}
}

\end{document}